%% file: main.tex
\documentclass[10pt,twocolumn,letterpaper]{article}

 \usepackage[pagenumbers]{cvpr} 


\definecolor{cvprblue}{rgb}{0.21,0.49,0.74}
\usepackage[pagebackref,breaklinks,colorlinks,allcolors=cvprblue]{hyperref}
\usepackage[linesnumbered,ruled,vlined]{algorithm2e} 
\usepackage{algpseudocode}
\usepackage{amssymb}  
 \usepackage{multirow}
 \usepackage{bm}  
 \usepackage{rotating}
 \usepackage{bbm}
 \usepackage{url}
  \usepackage{setspace}
 \usepackage{color}
\def\paperID{4864} 
\def\confName{CVPR}
\def\confYear{2025}

\title{Minding Fuzzy Regions: A Data-driven Alternating Learning Paradigm for Stable Lesion Segmentation}

\author{Lexin Fang\textsuperscript{1}, Yunyang Xu\textsuperscript{1}, Xiang Ma\textsuperscript{1}, Xuemei Li\textsuperscript{1,}\thanks{Corresponding author} , Caiming Zhang\textsuperscript{1}\\
\textsuperscript{1} School of Software, Shandong University\\
{\tt\small \{fanglexin, xuyunyang\}@mail.sdu.edu.cn, \{xiangma, xmli, czhang\}@sdu.edu.cn}
}

\begin{document}
\maketitle

\input{flx/Abstract}
\input{flx/Introduction}
\input{flx/RelatedWork}
\input{flx/Method}

\input{flx/Experiments}
\input{flx/Conclusion}
\section*{Acknowledgements}
\small
This work was supported by the Joint Fund of the National Natural Science Foundation of China under Grant Nos. U22A2033, U24A20219.
{
    \small
    
    \bibliographystyle{ieeenat_fullname}
    \bibliography{main}
}


\end{document}

%% file: flx/Abstract.tex
\begin{abstract}
Deep learning has achieved significant advancements in medical image segmentation, but existing models still face challenges in accurately segmenting lesion regions. The main reason is that some lesion regions in medical images have unclear boundaries, irregular shapes, and small tissue density differences, leading to label ambiguity. However, the existing model treats all data equally without taking quality differences into account in the training process, resulting in noisy labels negatively impacting model training and unstable feature representations. In this paper, a  \textbf{d}ata-driven \textbf{a}lternating \textbf{le}arning (\textbf{DALE}) paradigm is proposed to optimize the model's training process, achieving stable and high-precision segmentation. The paradigm focuses on two key points: (1) reducing the impact of noisy labels, and (2) calibrating unstable representations. To mitigate the negative impact of noisy labels, a loss consistency-based collaborative optimization method is proposed, and its effectiveness is theoretically demonstrated. Specifically, the label confidence parameters are introduced to dynamically adjust the influence of labels of different confidence levels during model training, thus reducing the influence of noise labels. To calibrate the learning bias of unstable representations, a distribution alignment method is proposed. This method restores the underlying distribution of unstable representations, thereby enhancing the discriminative capability of fuzzy region representations. Extensive experiments on various benchmarks and model backbones demonstrate the superiority of the DALE paradigm, achieving an average performance improvement of up to 7.16\%.
\end{abstract}

%% file: flx/Introduction.tex
\section{Introduction}
\label{sec:intro}
Medical image segmentation is a crucial step in computer-aided diagnosis (CAD), assisting doctors in identifying and locating internal organs and lesions. 
Recently, several deep learning-based lesion segmentation methods \cite{rahman2024emcad, chen2022novel, li2024lvit,liu2024swin} have significantly advanced the field, including expanding receptive fields \cite{chen2022novel, cao2021dilated}, introducing attention mechanisms \cite{rahman2024emcad, chen2022aau}, and incorporating additional modality information \cite{li2024lvit, zhong2023ariadne}.
\begin{figure}[t]
	\centering
	\includegraphics[width=0.95\linewidth]{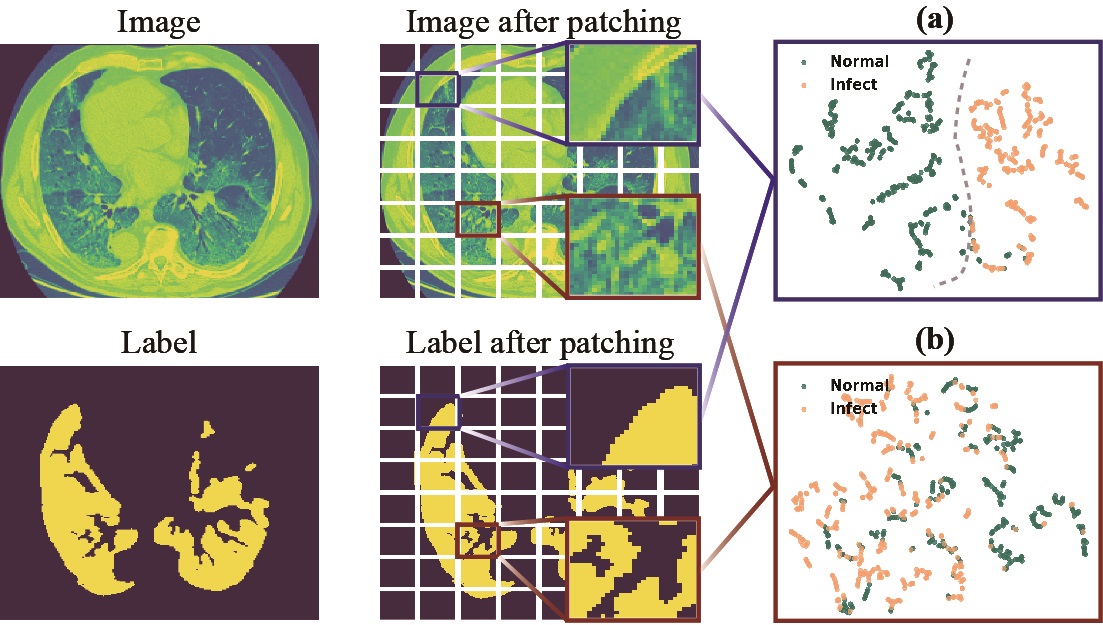}
	
	\caption{T-SNE visualization of features from fuzzy and non-fuzzy regions on the MosMedData+ dataset.
		(a) shows that model \cite{zhong2023ariadne} learns well-discriminative features in non-fuzzy regions (blue box).
		(b) shows significant overlap between lesion and normal tissue representations in fuzzy regions (red box).}
	\label{fig:1}
\end{figure}

In medical images, the regions of lesions appear unclear or irregular due to slight tissue density differences and irregular shapes. We refer to these regions as fuzzy regions, while others are non-fuzzy regions.
Fuzzy regions often contain noisy labels, which arise from unclear boundaries, as well as subjective differences, fatigue, and experience limitations among medical image labelers.
However, current segmentation models \cite{rahman2024emcad, li2024lvit,wang2023xbound,liu2024swin} treat all data equally during training, including data with noisy labels. Two key issues arise from this. First, noisy labels interfere with the model's training direction, misguiding the optimization process and causing error accumulation \cite{liu2020early}.
Second, the fuzzy region representation is unstable, which makes learning discriminative features challenging. This instability arises due to the unclear nature of fuzzy regions and low label confidence, causing the model to struggle in forming consistent feature representations.
As shown in \cref{fig:1}(b), in fuzzy regions, the feature projections of normal regions (green points) and infected regions (orange points) are mixed, indicating that different classes cannot form distinct class  boundaries. Considering the above reasons, existing methods have failed to segment fuzzy regions with high precision.

In order to overcome the challenge of model feature learning in fuzzy regions, we propose a novel \textbf{d}ata-driven \textbf{a}lternating \textbf{le}arning (\textbf{DALE}) paradigm based on \textbf{non-fuzzy} and \textbf{fuzzy} regions. This paradigm avoids the over-memorization \cite{liu2020early} of noisy labels and unstable features that can occur during traditional model training, improving segmentation accuracy and stability. Inspired by the process of human learning, the core idea is based on understanding \textbf{simple knowledge} before progressing to \textbf{more complex concepts}. 
In this way, the proposed DALE paradigm first utilizes reliable non-fuzzy region sets (non-fuzzy R-sets) for model training to learn stable feature representations, thereby providing prior knowledge and effective guidance for subsequent learning of fuzzy region sets (fuzzy R-sets).
During the training of fuzzy R-sets, confidence parameters which measure data label reliability are introduced to dynamically adjust the influence of different labels on model training.
Therefore, a loss consistency-based collaborative optimization method is designed to automatically learn the optimal confidence parameters. This method reduces the impact of noisy labels on training and improves the model's ability to learn effectively from high-confidence samples.
Moreover, we propose an unstable representation calibration method, which aligns unstable features in fuzzy regions with stable representation distributions. This process restores their underlying distributions and assists the model in learning more discriminative and stable feature representations.
DALE has been evaluated across multiple datasets and models, with significant improvements in the model's performance.

Our main contributions are the following:
\begin{itemize}
	\item A novel DALE paradigm is developed for training deep learning-based medical image segmentation models. This paradigm guides the training of fuzzy R-sets using reliable non-fuzzy R-sets, thus reducing the difficulty of model training.
	\item  A soft thresholding method is proposed to partition the dataset into fuzzy R-sets and non-fuzzy R-sets, taking into account the image complexity and irregularity of lesion boundaries.
	\item A collaborative optimization method based on loss consistency is proposed to learn and optimize the confidence parameters for labels. It is theoretically demonstrated that the method can learn the optimal confidence parameters, enabling the model to rely more on high-confidence samples and less on low-confidence samples during training.
	\item A distributed alignment method is designed to calibrate the unstable representation in the fuzzy region, enhancing the model's stability in lesion segmentation.
\end{itemize}

%% file: flx/RelatedWork.tex
\section{Related Work}
\subsection{Lesion Segmentation in Medical Imaging}
Lesion segmentation has been improved with various model improvements \cite{rahman2024emcad, chen2022novel, cao2021dilated, chen2022aau, li2024lvit, zhong2023ariadne} , including expanding the receptive field \cite{chen2022novel, cao2021dilated}, introducing attention mechanisms \cite{rahman2024emcad, chen2022aau}, and integrating multimodal information \cite{li2024lvit, zhong2023ariadne}.
To capture lesion features across multiple scales comprehensively, researchers often expand the receptive field to accommodate the diversity in lesion size and shape. For instance, D2U-Net \cite{cao2021dilated} expands the receptive field by introducing hybrid expansion convolution, which enables the model to more fully understand the overall morphology of the lesion. 
Considering the complexity of lesion images, 
EMCAD \cite{rahman2024emcad} proposes a new efficient multi-scale convolutional attention decoder that can capture complex spatial relationships and focus on key areas, improving segmentation performance for complex structural lesions. In addition to the variable size and shape of the lesion, the location of the lesion is difficult to detect. LViT model \cite{li2024lvit} proposes using the location and number of lesions in medical reports to guide the model in quickly locating the lesion area, helping the model better understand complex medical images. Although the above models are quite mature in design, they still face challenges in high-precision segmentation of fuzzy regions in lesions, mainly because these models do not fully consider the influence of noisy labels.
\begin{figure*}[t!]
	\centering
	\includegraphics[width=0.99\linewidth]{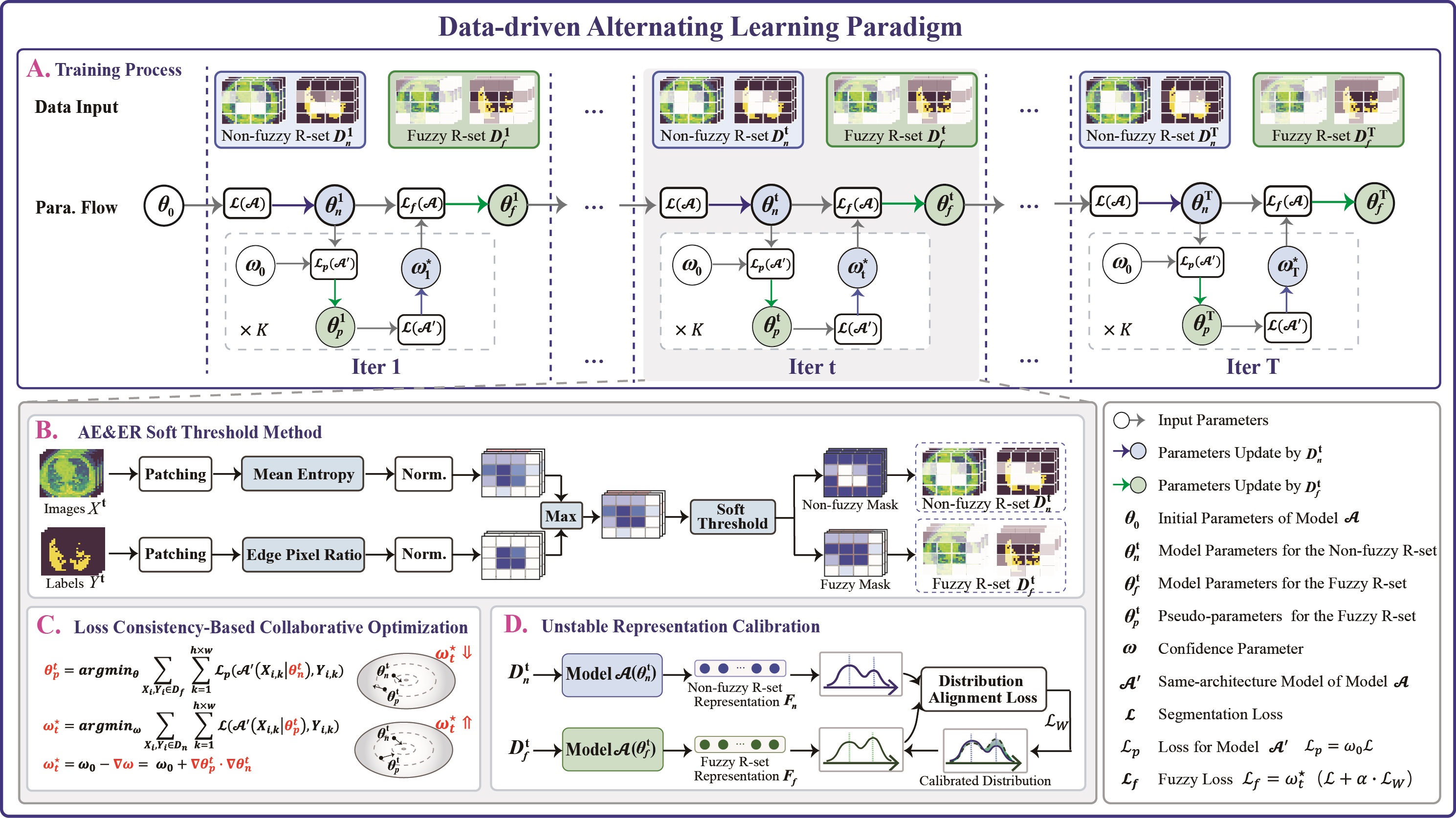}
	
	\caption{Training procedure of DALE.}
	\label{fig:2}
\end{figure*}
\subsection{Robust Learning with Noisy Labels}
It has become a consensus that deep models generally lead to better performance. However, the over-fitting ability of the deep model to noisy data \cite{jiang2018mentornet} may cause its stability to decline. To cope with the impact of label noise on deep models, some studies began to explore methods such as robust regularization \cite{wang2017multiclass, liu2020early}, label refurbishing \cite{song2019selfie, liu2022adaptive}, and co-teaching \cite{yu2019does, rong2023boundary}. For robust regularization, 
Liu et al. \cite{liu2020early} design a gradient between a regularized term and a mislabeled sample, implicitly preventing mislabeled memory. 
However, robust regularization methods rely on accurate label confidence measures, which can lead to cumulative error problems in the case of inaccurate label selection. The label correction method aims to generate more accurate pseudo-labels to replace the original noisy labels. 
For instance, SELFIE \cite{song2019selfie} sets a threshold to classify low-loss samples as clean samples and identifies noisy labels based on the fluctuation of sample predictions. 
However, these label refurbishing methods are limited by the early performance of the model. In order to avoid error correction, the Co-teaching method \cite{yu2019does} proposes to train two classifiers and optimize only the samples with divergent predictions. In light of this, BECO \cite{rong2023boundary} builds two parallel deep networks that are semantically segmenced through collaborative training, aiming to learn from each other and process all possible noisy data. 

%% file: flx/Method.tex
\section{Method}
\label{sec:Method}
We propose a \textbf{d}ata-driven \textbf{a}lternate \textbf{le}arning (\textbf{DALE}) paradigm to optimize the model's training process for stable lesion segmentation. DALE reduces errors in boundary segmentation caused by unclear lesion boundaries, irregular lesion shapes, and noisy labels, thereby enhancing the model's accuracy.
As the core of this paradigm, the alternating learning strategy utilizes the high confidence labels of non-fuzzy regions in the image to prioritize learning stable representations, thereby providing valuable prior knowledge and effective guidance for the subsequent learning of representations in the fuzzy regions. 

Part A of \cref{fig:2} illustrates the training process of this paradigm, with the $t$-th iteration shown in the middle part.
Fuzzy R-sets $D_f^t$ and non-fuzzy R-sets $D_n^t$ are created first using the average entropy and edge ratio (AE\&ER) soft threshold method (Part B of \cref{fig:2}).
Then, the model $\mathcal{A}$ is trained using the non-fuzzy R-set $D_n^t$ and the commonly used segmentation loss function $\mathcal{L}$ for backpropagation, obtaining the model parameters $\theta_n^t$. 
Based on the parameters $\theta_n^t$, fuzzy R-set $D_f^t$ are used to train the model, and backpropagation is performed using the fuzzy loss function $\mathcal{L}_f$ to obtain the parameters $\theta_f^t$.
Following the above process, the model $\mathcal{A}$ updates its parameters through alternating learning of $\theta_n^t$ and $\theta_f^t$.
The fuzzy loss function $\mathcal{L}_f$ (\cref{eq.15}) consists of the loss function $\mathcal{L}$, the label confidence parameter $\omega_t^{\star}$  (Part C of \cref{fig:2}), and the distribution alignment loss $\mathcal{L}_W$ (Part D of \cref{fig:2}).
When the label's confidence is low, the corresponding $\omega$ value decreases. In this way, the model gives less weight to these samples during training, reducing the influence of the low confidence label on the training process effectively.
 $\mathcal{L}_W$ is used to calibrate the unstable representations in fuzzy regions, enhancing the model's ability to learn discriminative features from those regions.
 
The alternating learning process is as follows,
\begin{equation}
	\theta_n^t=argmin_\theta\sum_{X_i,Y_i\in D_n^t}\sum_{k=1}^{h\times w}\mathcal{L}(\mathcal{A}(X_{i,k}|\theta_f^{t-1}),Y_{i,k}),\label{eq.1}
\end{equation}
\begin{equation}
	\theta_f^t=argmin_\theta\sum_{X_i,Y_i\in D_f^t}\sum_{k=1}^{h\times w}\mathcal{L}_f(\mathcal{A}(X_{i,k}|\theta_n^t),Y_{i,k})\label{eq.2},
\end{equation}
when $t=1$, $\theta_f^{t-1}=\theta_0$, which is the model initialization parameter. $\mathcal{L}_f$ is defined in \cref{eq.15}.

The AE\&ER soft threshold method is detailed in \cref{sec:3.1}. 
The loss consistency-based collaborative optimization method is described in \cref{sec:3.2}.
The unstable representation calibration is explained in \cref{sec:3.3}.

\subsection{AE\&ER Soft Threshold Method}
\label{sec:3.1}
There are various levels of fuzziness in image patches, as illustrated by the red and blue boxes in \cref{fig:1}, which affect the model's ability to learn features. Furthermore, some patches contain complex, irregular lesion edges, which tend to produce noisy labels. Therefore, we propose a soft threshold method that divides an image into fuzzy R-set and non-fuzzy R-set based on the degree of fuzziness and edge complexity of image patches.

Specifically, given a training dataset $D=\{(X_i,Y_i)\}_{i=1}^N$, where $X_i\in R^{H\times W\times3}$ represents the training image, $Y_i\in R^{H\times W}$ denotes the corresponding label. First, each sample $(X_i, Y_i)$ is partitioned, with the image $X_i$ and the corresponding label $Y_i$ divided into $L$ patches of size $h \times w$, respectively.
The fuzziness of the sample $(X_i, Y_i)$ is defined as $R_i = \{r_{i,j}\}_{j=1}^L$, which is calculated by the average entropy of each patch in the image $X_i$. The average entropy (AE) $r_{i,j}$ of the $j$-th patch is computed as,
\begin{equation}r_{i,j}=-\frac1{h\times w}\sum_{k=1}^{h\times w}p_{i,j,k}log(p_{i,j,k}),\label{eq.3}\end{equation}
where $p_{i,j,k}$ represents the value probability of the $k$-th pixel in the $j$-th patch of the image $X_i$. The higher the average entropy value $r_{i,j}$, the more uniform the pixel distribution in the $j$-th patch.
In addition, the edge complexity of the sample $(X_i, Y_i)$ is defined as $E_i = \{e_{i,j}\}_{j=1}^L$, which is calculated by the ratio of edge pixels in each patch of the label $Y_i$.
Regions with a higher proportion of edge pixels have more irregular edge structures and are more  prone to noisy labels in general. The edge pixel ratio (ER) of the $j$-th patch is calculated as,
\begin{equation}e_{i,j}=\frac{\Sigma_{k=1}^{h\times w}\mathbbm{l}[I_{i,j,k}^e=1]}{h\times w},\label{eq.4}\end{equation}
where $\mathbbm{l}[\cdot]$ is the indicator function, and $I_i^e$ is the edge map generated from the label $Y_i$. $I_{i,j,k}^e = 1$ indicates that the $k$-th pixel in the $j$-th patch of label $Y_i$ is an edge pixel.
Next, the fuzziness and edge complexity are normalized separately (as shown in \cref{fig:2} as ``Norm.'') as follows,
\begin{equation}\small r_{i,j}^{std}=\frac{r_{i,j}-\min(R_i)}{\max(R_i)-\min(R_i)},\quad e_{i,j}^{std}=\frac{e_{i,j}-\min(E_i)}{\max(E_i)-\min(E_i)}.\label{eq.5}\end{equation}

Then, soft threshold masking is applied to the sample $(X_i, Y_i)$. Compared to using image patches or hard threshold masking directly, soft thresholding of samples can preserve important spatial relationships without harming the overall spatial structure.
Specifically, set the soft threshold $\tau$, and obtain the fuzzy mask $M_i^f=\{M_{i,j}^f\}_{j=1}^L$ and the non-fuzzy mask $M_i^n=\{M_{i,j}^n\}_{j=1}^L$,
\begin{equation}\small\begin{gathered}M_{i,j}^{f}=\begin{cases}1,if\quad m_{i,j}>\tau,\\m_{i,j},else.\end{cases} M_{i,j}^{n}=\begin{cases}1,if\quad m_{i,j}<\tau,\\\tau- m_{i,j},else.\end{cases}\end{gathered}\label{eq.6}
\end{equation}
where $\tau$ is the hyper-parameter. To account for both image fuzziness and edge information, $m_{i,j}$ is defined as $m_{i,j} = \max(r_{i,j}^{std}, e_{i,j}^{std})$.
Using the mask $M_i^f$ and $M_i^n$, the training set $D$ is divided into a fuzzy R-set $ D_{f}=\{(X_{i}^{f},Y_{i}^{f})\}_{i=1}^{N}$ and a non-fuzzy R-set $D_n=\{(X_i^n,Y_i^n)\}_{i=1}^N$,
\begin{equation}
\begin{aligned}
	X_i^f=M_i^f\odot X_i, X_i^n=M_i^n\odot X_i; \\
	Y_i^f=M_i^f\odot Y_i, Y_i^n=M_i^n\odot Y_i.
\end{aligned}\label{eq.7}
\end{equation}

\subsection{Loss Consistency-Based Collaborative Optimization}
\label{sec:3.2}
To detect and mitigate the impact of noisy data in the fuzzy R-set on the model, label confidence parameters, denoted as $\omega$, are introduced during the training process with the fuzzy R-set (\cref{eq.15}). 
When a sample in the fuzzy set is detected as noisy, its corresponding confidence parameter $\omega$ is reduced, which makes the model rely less on low confidence labels during training.
Therefore, estimating parameter $\omega$ accurately is crucial. 

Existing noise label processing methods \cite{song2019selfie} typically rely on manually set thresholds to identify noisy labels, which may lead to errors. From an optimization perspective, dynamically adapting label confidence can yield more accurate results when leveraging the bi-level minimization optimization strategy of meta-learning \cite{wu2021learning, pham2021meta, zhao2024stable}. Accordingly, we introduce the following proposition to achieve collaborative optimization.

\noindent\textbf{Proposition:} The proposed collaborative optimization method based on loss consistency (\cref{eq:8,eq:9}) can determine the optimal confidence parameter \( \omega_t^{\star} \). In this way, if the labels of the samples in the fuzzy R-set are correct, the optimization process will reduce the loss in both the fuzzy and non-fuzzy R-set simultaneously during training. 
 Let model \( \mathcal{A} \) and its shadow model \( \mathcal{A}^{'} \), which have the same structure but different parameters, be given. First, the model \( \mathcal{A}^{'}, \) is trained on the fuzzy R-sets \( D_f^t \) using the \( \theta_n^t \) obtained from training on the non-fuzzy R-sets \( D_n^t \), resulting in the pseudo-parameters \( \theta_p^t \),
\begin{equation}
	\theta_p^t = \arg \min_{\theta} \sum_{(X_i, Y_i) \in D_f^t} \sum_{k=1}^{h \times w} \mathcal{L}_p \left( \mathcal{A}^{'} \left( X_{(i,k)} | \theta_n^t \right), Y_{(i,k)} \right) .
	 \label{eq:8}
\end{equation}
Then, the confidence parameter $\omega$ is optimized using the following equation,
\begin{equation}
	\omega_t^{\star} = \arg \min_{\omega} \sum_{(X_i, Y_i) \in D_n^t} \sum_{k=1}^{h \times w} \mathcal{L} \left( \mathcal{A}^{'} \left( X_{(i,k)} | \theta_p^t \right), Y_{(i,k)} \right) ,
	\label{eq:9}
\end{equation}
where $\mathcal{L}_p=\omega_0\mathcal{L}$, \( \omega_0 \in \mathbb{R}^{H \times W} \) is initialized as a matrix of ones. When the label of a sample is correct, \( \omega_t^{\star} \) increases; otherwise, it decreases.

\noindent\textbf{Proof:} In gradient descent optimization, the following equations can be derived,
\begin{equation}
\omega_{t}^{\star}=\omega_{0}-\sum_{X_{i}\in D_{n}^t}\frac{\mathcal{L}(X_{i,k}|\theta_{p}^{t})}{\partial\omega}|_{{\omega}_{0}}.
\label{eq.10}\end{equation}
Expanding the partial derivative yields,
\begin{equation}
\begin{aligned}
&\omega_{t}^{\star}=\omega_0-\sum_{X_i\in D_n^t}\frac{\mathcal{L}(X_{i,k}|\theta_p^t)}{\partial\theta}|_{{\theta}_p^t}\cdot\frac{\partial\theta_p^t}{\partial\omega}|_{{\omega}_0}\\&
=\omega_0+\sum_{X_i\in D_n^t}\frac{\mathcal{L}(X_{i,k}|\theta_p^t)}{\partial\theta}|_{\theta_p^t}\cdot\sum_{X_i\in D_f^t}\frac{\mathcal{L}(X_{i,k}|\theta_n^t)}{\partial\theta}|_{{\theta}_n^t}.
\end{aligned}\label{eq.11}
\end{equation}
It can be seen from \cref{eq.11} that when the sample has similar gradient direction in the optimization process of fuzzy R-set parameters $\theta_n^t$ and non-fuzzy R-set parameters $\theta_p^t$, the confidence $\omega$ of the sample will increase, thereby enhancing the influence of high-confidence samples on the model's training. Conversely, if the gradient is in the opposite direction, it indicates that the sample contains a noisy label,  causing the confidence parameter $\omega$ to decrease, thereby reducing the impact of noisy samples on the model's training. 

The optimization process of the confidence parameter $\omega$ can be repeated $K$ times. The final optimized confidence parameter $\omega_t^{\star}$ is then applied in \cref{eq.15} to calculate the fuzzy loss, ensuring dynamic adjustment of the influence of labels with different levels of confidence.

\subsection{Unstable Representation Calibration}
\label{sec:3.3}
\begin{figure}[t]
	\centering
	\includegraphics[width=0.99\linewidth]{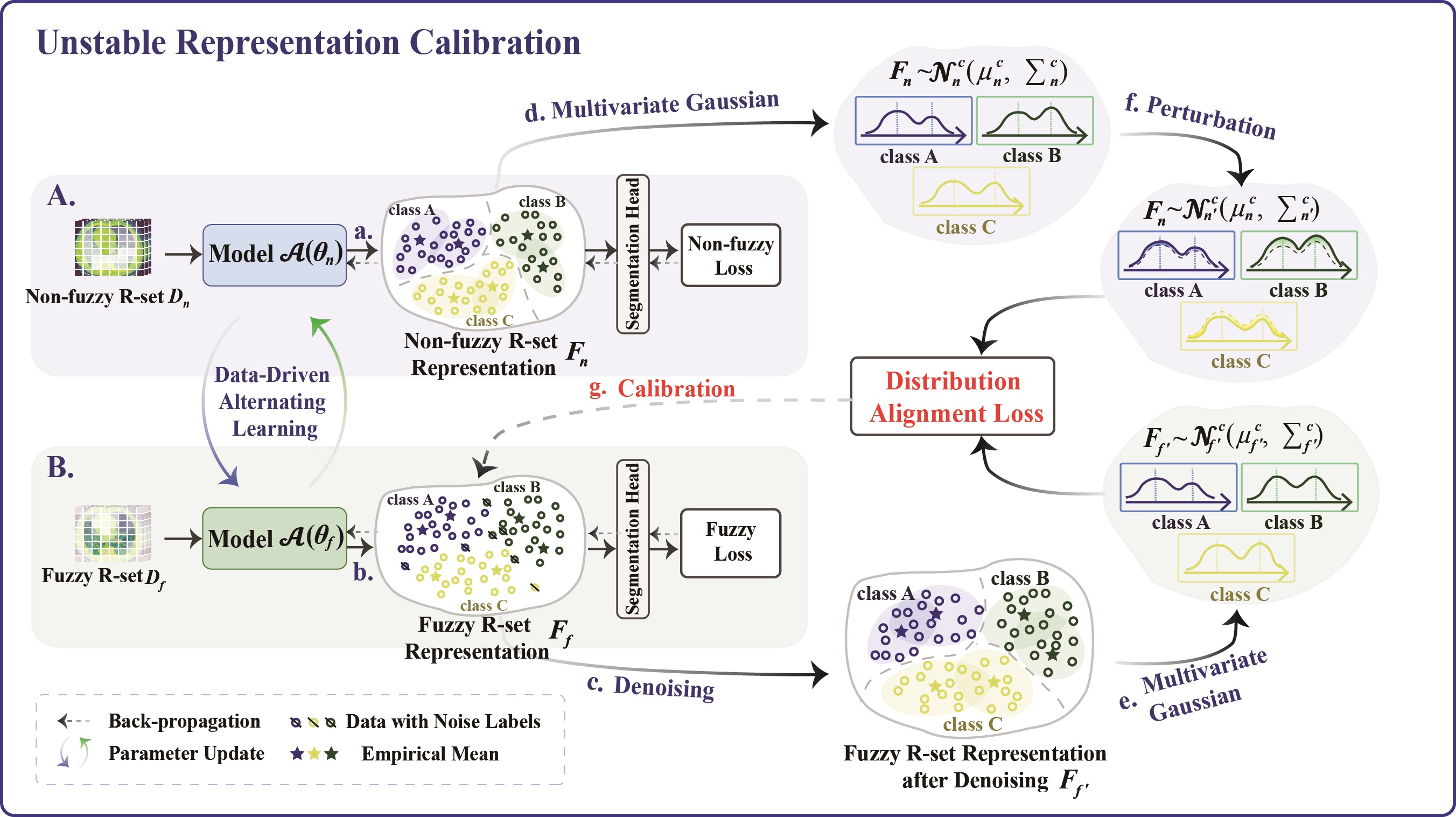}
	
	\caption{Unstable representation calibration. Calibration is performed in each iteration, with iteration superscripts omitted for simplicity.}
	\label{fig:3}
\end{figure}
Fuzzy regions can comprise unclear boundaries and shapes, and noise labels may be present, resulting in poor representation quality for the model. By contrast, non-fuzzy R-sets are more structured and contrasted. 
Using non-fuzzy R-sets for training the model helps to extract more discriminative representations (see \cref{fig:1}). In light of this, we propose calibrating the unstable representations in fuzzy R-sets using the representation distribution extracted from non-fuzzy R-sets, as shown in \cref{fig:3}.

In the absence of noisy labels, it is typically assumed that the data representations of each class follow a multivariate Gaussian distribution \cite{zhang2023noisy, wang2024label, zhang2022tackling}. Feature representations \( F_n \) and \( F_f \) are extracted from the training phases of the non-fuzzy R-set (\cref{eq.1}) and the fuzzy R-set (\cref{eq.2}) before entering the model’s segmentation head, as shown in operations a and b in \cref{fig:3}. To avoid interference from noisy labels, we remove the corresponding part of the noise data in the representations \( F_f \), and obtain the denoised representation \( F_f' \) (operation c in \cref{fig:3}). Noise data can be detected by judging the gradient direction of \( \omega_t^\star \) (\cref{eq.11}). If the gradient of \( \omega_t^\star \) is negative, it indicates that the corresponding region likely contains noisy labels,
\begin{equation}
	F_{f'} = F_f \cdot \mathbbm{l}[\nabla \omega > 0], \label{eq.12}
\end{equation}
where \( \nabla \omega \) is the gradient of \( \omega_t^* \). \( \mathbbm{l}[\cdot] \) is the indicator function applied to retain the non-noise portion of the representation.

Next, we estimate the multivariate Gaussian distribution $\mathcal{N}_n^c(\mu_n^c,\Sigma_n^c)$ (see operation d in \cref{fig:3}) for each class of the non-fuzzy R-set representations \( F_n \), and the multivariate Gaussian distribution $\mathcal{N}_{f^{\prime}}^{c}(\mu_{f^{\prime}}^{c},\Sigma_{f^{\prime}}^{c})$ (see operation e in Figure 3) for the denoised fuzzy R-set representations \( F_{f'} \), where \( c \in C \) and \( C \) is the number of classes. It is important to note that the multivariate Gaussian distribution for each class derived from the non-fuzzy R-set may not cover all data for that class, causing a shift in the distribution, i.e., covariate shift.  It can be alleviated by adding an appropriate perturbation term to the estimated empirical covariance matrix (operation f in \cref{fig:3}). For class \( c \), the covariance update is,
\begin{equation}\Sigma_{n^{\prime}}^c=\Sigma_n^c+\varepsilon\mathbf{1},\label{eq.13}\end{equation}
where \( \mathbf{1} \) represents a matrix of ones, and \( \varepsilon \) is a random number that controls the degree of perturbation. The updated distribution \( N_{n'}^c(\mu_n^c, \Sigma_{n'}^c) \) of \( F_n \) is then obtained.

We define the Wasserstein distance \cite{rubner2000earth, givens1984class} between two multivariate Gaussian distributions to measure the difference in representation distributions between fuzzy and non-fuzzy R-sets. Wasserstein distance is used instead of KL divergence because it remains stable even when there is no overlap between distributions.
The distribution alignment loss $\mathcal{L}_{W}$ is as follows,
\begin{equation}
	\small
	\begin{aligned}&\mathcal{L}_{W}=\sum_{c\in C}D_W^c(\mathcal{N}_{n^{\prime}}^c(\mu_n^c,\Sigma_{n^{\prime}}^c),\mathcal{N}_{f^{\prime}}^c(\mu_{f^{\prime}}^c,\Sigma_{f^{\prime}}^c))\\&=\sum_{c\in C}(||\mu_n^c-\mu_{f^{\prime}}^c||^2+tr(\Sigma_{n^{\prime}}^c+\Sigma_{f^{\prime}}^c-2(\sqrt{\Sigma_{n^{\prime}}^c}\Sigma_{f^{\prime}}^c\sqrt{\Sigma_{n^{\prime}}^c})^{\frac12})),\end{aligned}\label{eq.14}\end{equation}
here, $tr(\cdot)$ denotes the trace of a matrix.

In the process of model optimization, the fuzzy loss function $\mathcal{L}_{f}$ introduces the distribution alignment loss $\mathcal{L}_{W}$ on the basis of the commonly used segmentation loss $\mathcal{L}$, so that the unstable feature distribution of fuzzy R-set can be aligned with the distribution of non-fuzzy R-set as much as possible (operation g in \cref{fig:3}), so as to restore the stable feature representation. The fuzzy loss function $\mathcal{L}_f$ is,
\begin{equation}\mathcal{L}_f=\omega_t^\star(\mathcal{L}+\alpha\cdot\mathcal{L}_W),
\label{eq.15}
\end{equation}
where $\alpha$ is a hyper-parameter to balance the $\mathcal{L}_{W}$ and the $\mathcal{L}$.

The DALE process is shown in Algorithm 1.
\begin{algorithm}
\small
\caption{Data-driven Alternating Learning Paradigm}
\label{al1}
\KwIn{Model $\mathcal{A}$, training dataset $D$, initial model parameters $\theta_0$, maximum number of iterations $T$, number of iterations for $\omega$ optimization $K$.}
\KwOut{Well-trained model $\mathcal{A}$ parameters $\theta_f^T$.}
Initialize model $\mathcal{A}$ with $\theta_0$\;
\For{$t \leftarrow 1$ \KwTo $T$}{
	Split $D$ into $D_n^t$ and $D_f^t$ by \cref{eq.3,eq.4,eq.5,eq.6,eq.7}\;
	Extract $\mathcal{F}_n^t$ from model $\mathcal{A}$ using input $D_n^t$\;
	Optimize $\theta_n^t$ via $D_n^t$ by \cref{eq.1}\;
	Initialize the model \( \mathcal{A}' \) with \( \theta_n^t \), identical to model \( \mathcal{A} \)\;
	\For{$k \leftarrow 1$ \KwTo $K$}{
		Optimize $\theta_p^t$ on the model \( \mathcal{A}' \) via $D_f^t$ by \cref{eq:8}\;
		Optimize $\omega_t$ on the model \( \mathcal{A}' \) via $D_n^t$ by \cref{eq:9}\;
	}
	Obtain the optimal \( \omega_t^{\star} \)\;
	Extract $\mathcal{F}_f^t$ from model $\mathcal{A}$ using input $D_f^t$\;
	Get the $\mathcal{L}_W$  by \cref{eq.12,eq.13,eq.14}\;
	Obtain fuzzy loss $\mathcal{L}_f$ by \cref{eq.15}\;
	Optimize $\theta_f^t$ via $D_f^t$ by \cref{eq.2}\;
}
Return $\theta_f^T$;
	
\end{algorithm}

%% file: flx/Experiments.tex
\newcommand{\mypara}[1]{{\textbf{#1\;\;}}}
\begin{table}[b]
	\small
	
	\scalebox{0.73}{
		\begin{tabular}{@{}ccccccc@{}}
			\toprule
			\multicolumn{1}{c|}{}                                  & \multicolumn{1}{c|}{}                                                                                      & \multicolumn{4}{c|}{\textbf{MosMedData+}}                                                                                      &                                                          \\ \cmidrule(lr){3-6}
			\multicolumn{1}{c|}{\multirow{-2}{*}{\textbf{Model}}} & \multicolumn{1}{c|}{\multirow{-2}{*}{\textbf{Model Type}}}                                                   & \textbf{Dice$\uparrow$} & \textbf{mIoU$\uparrow$} & \textbf{95HD$\downarrow$} & \multicolumn{1}{c|}{\textbf{ASD$\downarrow$}} &\multirow{-2}{*}{\textbf{\begin{tabular}[c]{@{}c@{}}Growth \\ Rate\end{tabular}}}              \\ \midrule
			\multicolumn{1}{c|}{Unet}                              & \multicolumn{1}{c|}{}                                                                                      & 0.646                   & 0.507                   & 25.428                    & \multicolumn{1}{c|}{5.826}                    & {\color[HTML]{FE0000} }                                  \\
			\multicolumn{1}{c|}{\textbf{*Unet}}                   & \multicolumn{1}{c|}{}                                                                                      & \textbf{0.694}          & \textbf{0.554}          & \textbf{25.250}           & \multicolumn{1}{c|}{\textbf{5.320}}           & \multirow{-2}{*}{{\color[HTML]{FE0000} \textbf{+7.45\%}}} \\ \cmidrule(r){1-1} \cmidrule(l){3-7} 
			\multicolumn{1}{c|}{Unet++}                            & \multicolumn{1}{c|}{}                                                                                      & 0.716                   & 0.582                   & 22.109                    & \multicolumn{1}{c|}{4.326}                    & {\color[HTML]{FE0000} }                                  \\
			\multicolumn{1}{c|}{\textbf{*Unet++}}                  & \multicolumn{1}{c|}{}                                                                                      & \textbf{0.741}          & \textbf{0.610}          & \textbf{3.595}            & \multicolumn{1}{c|}{\textbf{15.577}}          & \multirow{-2}{*}{{\color[HTML]{FE0000} \textbf{+3.42\%}}} \\ \cmidrule(r){1-1} \cmidrule(l){3-7} 
			\multicolumn{1}{c|}{EGEUNET}                           & \multicolumn{1}{c|}{}                                                                                      & 0.693                   & 0.558                   & 20.906                    & \multicolumn{1}{c|}{4.203}                    & {\color[HTML]{FE0000} }                                  \\
			\multicolumn{1}{c|}{\textbf{*EGEUNET}}                 & \multicolumn{1}{c|}{}                                                                                      & \textbf{0.736}          & \textbf{0.611}          & \textbf{17.296}           & \multicolumn{1}{c|}{\textbf{3.429}}           & \multirow{-2}{*}{{\color[HTML]{FE0000} \textbf{+6.16\%}}} \\ \cmidrule(r){1-1} \cmidrule(l){3-7} 
			\multicolumn{1}{c|}{EMCAD}                             & \multicolumn{1}{c|}{}                                                                                      & 0.686                   & 0.552                   & 21.751                    & \multicolumn{1}{c|}{4.807}                    & {\color[HTML]{FE0000} }                                  \\
			\multicolumn{1}{c|}{\textbf{*EMCAD}}                   & \multicolumn{1}{c|}{\multirow{-8}{*}{\textbf{CNN-based}}}                                                  & \textbf{0.721}          & \textbf{0.590}          & \textbf{19.577}           & \multicolumn{1}{c|}{\textbf{4.595}}           & \multirow{-2}{*}{{\color[HTML]{FE0000} \textbf{+5.03\%}}} \\ \midrule
			\multicolumn{1}{c|}{Convformer}                        & \multicolumn{1}{c|}{}                                                                                      & 0.681                   & 0.536                   & 23.250                    & \multicolumn{1}{c|}{4.800}                    & {\color[HTML]{FE0000} }                                  \\
			\multicolumn{1}{c|}{\textbf{*Convformer}}              & \multicolumn{1}{c|}{}                                                                                      & \textbf{0.716}          & \textbf{0.581}          & \textbf{21.231}           & \multicolumn{1}{c|}{\textbf{4.165}}           & \multirow{-2}{*}{{\color[HTML]{FE0000} \textbf{+5.11\%}}} \\ \cmidrule(r){1-1} \cmidrule(l){3-7} 
			\multicolumn{1}{c|}{TransFuse}                         & \multicolumn{1}{c|}{}                                                                                      & 0.714                   & 0.582                   & 21.536                    & \multicolumn{1}{c|}{4.740}                    & {\color[HTML]{FE0000} }                                  \\
			\multicolumn{1}{c|}{\textbf{*TransFuse}}               & \multicolumn{1}{c|}{}                                                                                      & \textbf{0.759}          & \textbf{0.612}          & \textbf{14.342}           & \multicolumn{1}{c|}{\textbf{2.325}}           & \multirow{-2}{*}{{\color[HTML]{FE0000} \textbf{+6.37\%}}} \\ \cmidrule(r){1-1} \cmidrule(l){3-7} 
			\multicolumn{1}{c|}{Xboundformer}                      & \multicolumn{1}{c|}{}                                                                                      & 0.679                   & 0.543                   & 26.056                    & \multicolumn{1}{c|}{6.257}                    & {\color[HTML]{FE0000} }                                  \\
			\multicolumn{1}{c|}{\textbf{*Xboundformer}}            & \multicolumn{1}{c|}{\multirow{-6}{*}{\textbf{\begin{tabular}[c]{@{}c@{}}CNN-\\ Transformer\\hybrid\end{tabular}}}} & \textbf{0.709}          & \textbf{0.572}          & \textbf{20.534}           & \multicolumn{1}{c|}{\textbf{4.674}}           & \multirow{-2}{*}{{\color[HTML]{FE0000} \textbf{+4.36\%}}} \\ \midrule
			\multicolumn{1}{c|}{SwinUmamba}                        & \multicolumn{1}{c|}{}                                                                                      & 0.701                   & 0.568                   & 22.576                    & \multicolumn{1}{c|}{5.160}                    & {\color[HTML]{FE0000} }                                  \\
			\multicolumn{1}{c|}{\textbf{*SwinUmamba}}              & \multicolumn{1}{c|}{\multirow{-2}{*}{\textbf{Mamba}}}                                                      & \textbf{0.729}          & \textbf{0.598}          & \textbf{19.327}           & \multicolumn{1}{c|}{\textbf{4.083}}           & \multirow{-2}{*}{{\color[HTML]{FE0000} \textbf{+3.92\%}}} \\ \midrule
			\multicolumn{1}{c|}{LanGuide}                          & \multicolumn{1}{c|}{}                                                                                      & 0.735                   & 0.612                   & 16.879                    & \multicolumn{1}{c|}{3.285}                    & {\color[HTML]{FE0000} }                                  \\
			\multicolumn{1}{c|}{\textbf{*LanGuide}}                & \multicolumn{1}{c|}{\multirow{-2}{*}{\textbf{Multimodal}}}                                                & \textbf{0.780}          & \textbf{0.639}          & \textbf{10.682}           & \multicolumn{1}{c|}{\textbf{1.879}}           & \multirow{-2}{*}{{\color[HTML]{FE0000} \textbf{+6.11\%}}} \\ \midrule
			\multicolumn{2}{c|}{\textbf{Average Growth Rate}}                                                                                                                    & \multicolumn{5}{c}{{\color[HTML]{FE0000} \textbf{+5.33\%}}}                                                                                                                               \\ \bottomrule
	\end{tabular}}
	\caption{Evaluation of the DALE on various advanced models using MosMedData+ dataset. * Indicates a model trained using DALE. Red indicates the growth rate of Dice.}
	\label{tab:1}
\end{table}
\begin{table*}[]
	\small
	\scalebox{0.69}{
		\begin{tabular}{@{}cc|cccc|c|ccccccccccc@{}}
			\toprule
			\multicolumn{1}{c|}{}                                  &                                                                                       & \multicolumn{4}{c|}{\textbf{ISIC2016\&ph2}}                                                              &                                                           & \multicolumn{2}{c|}{\textbf{EndoScene}}                                & \multicolumn{2}{c|}{\textbf{CVC-ClinicDB}}                             & \multicolumn{2}{c}{\textbf{CVC-ColonDB}}                               & \multicolumn{2}{c|}{\textbf{ETIS}}                                     & \multicolumn{2}{c|}{\textbf{Kvasir}}                                   & \multicolumn{1}{l}{}                                       \\ \cmidrule(lr){3-6} \cmidrule(lr){8-17}
			\multicolumn{1}{c|}{\multirow{-2}{*}{\textbf{Model}}} & \multirow{-2}{*}{\textbf{Model Type}}                                                   & \textbf{Dice$\uparrow$} & \textbf{mIoU$\uparrow$} & \textbf{95HD$\downarrow$} & \textbf{ASD$\downarrow$} &\multirow{-2}{*}{\textbf{\begin{tabular}[c]{@{}c@{}}Growth \\ Rate\end{tabular}}}               & \textbf{Dice$\uparrow$} & \multicolumn{1}{c|}{\textbf{mIoU$\uparrow$}} & \textbf{Dice$\uparrow$} & \multicolumn{1}{c|}{\textbf{mIoU$\uparrow$}} & \textbf{Dice$\uparrow$} & \multicolumn{1}{c|}{\textbf{mIoU$\uparrow$}} & \textbf{Dice$\uparrow$} & \multicolumn{1}{c|}{\textbf{mIoU$\uparrow$}} & \textbf{Dice$\uparrow$} & \multicolumn{1}{c|}{\textbf{mIoU$\uparrow$}} & \multirow{-2}{*}{\textbf{\begin{tabular}[c]{@{}c@{}}Growth \\ Rate\end{tabular}}} \\ \midrule
			\multicolumn{1}{c|}{Unet}                              &                                                                                       & 0.824                   & 0.735                   & 16.768                    & 4.455                    & {\color[HTML]{FE0000} }                                   & 0.794                   & \multicolumn{1}{c|}{0.692}                   & 0.818                   & \multicolumn{1}{c|}{0.747}                   & 0.752                   & \multicolumn{1}{c|}{0.658}                   & 0.574                   & \multicolumn{1}{c|}{0.493}                   & 0.818                   & \multicolumn{1}{c|}{0.746}                   & {\color[HTML]{FE0000} }                                    \\
			\multicolumn{1}{c|}{\textbf{*Unet}}                    &                                                                                       & \textbf{0.857}          & \textbf{0.759}          & \textbf{13.399}           & \textbf{2.645}           & \multirow{-2}{*}{{\color[HTML]{FE0000} \textbf{+4.07\%}}} & \textbf{0.863}          & \multicolumn{1}{c|}{\textbf{0.792}}          & \textbf{0.870}          & \multicolumn{1}{c|}{\textbf{0.793}}          & \textbf{0.779}          & \multicolumn{1}{c|}{\textbf{0.694}}          & \textbf{0.652}          & \multicolumn{1}{c|}{\textbf{0.567}}          & \textbf{0.849}          & \multicolumn{1}{c|}{\textbf{0.775}}          & \multirow{-2}{*}{{\color[HTML]{FE0000} \textbf{+6.81\%}}}  \\ \cmidrule(r){1-1} \cmidrule(l){3-18} 
			\multicolumn{1}{c|}{Unet++}                            &                                                                                       & 0.888                   & 0.809                   & 10.470                    & 1.685                    & {\color[HTML]{FE0000} }                                   & 0.865                   & \multicolumn{1}{c|}{0.787}                   & 0.830                   & \multicolumn{1}{c|}{0.764}                   & 0.721                   & \multicolumn{1}{c|}{0.635}                   & 0.653                   & \multicolumn{1}{c|}{0.571}                   & 0.873                   & \multicolumn{1}{c|}{0.799}                   & {\color[HTML]{FE0000} }                                    \\
			\multicolumn{1}{c|}{\textbf{*Unet++}}                  &                                                                                       & \textbf{0.912}          & \textbf{0.844}          & \textbf{5.451}            & \textbf{0.871}           & \multirow{-2}{*}{{\color[HTML]{FE0000} \textbf{+2.61\%}}} & \textbf{0.887}          & \multicolumn{1}{c|}{\textbf{0.811}}          & \textbf{0.901}          & \multicolumn{1}{c|}{\textbf{0.843}}          & \textbf{0.784}          & \multicolumn{1}{c|}{\textbf{0.696}}          & \textbf{0.709}          & \multicolumn{1}{c|}{\textbf{0.628}}          & \textbf{0.903}          & \multicolumn{1}{c|}{\textbf{0.839}}          & \multirow{-2}{*}{{\color[HTML]{FE0000} \textbf{+6.15\%}}}  \\ \cmidrule(r){1-1} \cmidrule(l){3-18} 
			\multicolumn{1}{c|}{EMCAD}                             &                                                                                       & 0.919                   & 0.857                   & 5.105                     & 0.882                    & {\color[HTML]{FE0000} }                                   & 0.852                   & \multicolumn{1}{c|}{0.769}                   & 0.837                   & \multicolumn{1}{c|}{0.765}                   & 0.771                   & \multicolumn{1}{c|}{0.681}                   & 0.626                   & \multicolumn{1}{c|}{0.545}                   & 0.869                   & \multicolumn{1}{c|}{0.796}                   & {\color[HTML]{FE0000} }                                    \\
			\multicolumn{1}{c|}{\textbf{*EMCAD}}                   &                                                                                       & \textbf{0.936}          & \textbf{0.883}          & \textbf{3.003}            & \textbf{0.549}           & \multirow{-2}{*}{{\color[HTML]{FE0000} \textbf{+1.83\%}}} & \textbf{0.879}          & \multicolumn{1}{c|}{\textbf{0.804}}          & \textbf{0.909}          & \multicolumn{1}{c|}{\textbf{0.853}}          & \textbf{0.784}          & \multicolumn{1}{c|}{\textbf{0.696}}          & \textbf{0.788}          & \multicolumn{1}{c|}{\textbf{0.703}}          & \textbf{0.913}          & \multicolumn{1}{c|}{\textbf{0.856}}          & \multirow{-2}{*}{{\color[HTML]{FE0000} \textbf{+8.05\%}}}  \\ \cmidrule(r){1-1} \cmidrule(l){3-18} 
			\multicolumn{1}{c|}{EGEUNET}                           &                                                                                       & 0.894                   & 0.815                   & 7.548                     & 1.241                    & {\color[HTML]{FE0000} }                                   & 0.853                   & \multicolumn{1}{c|}{0.779}                   & 0.863                   & \multicolumn{1}{c|}{0.792}                   & 0.728                   & \multicolumn{1}{c|}{0.639}                   & 0.619                   & \multicolumn{1}{c|}{0.546}                   & 0.853                   & \multicolumn{1}{c|}{0.779}                   & {\color[HTML]{FE0000} }                                    \\
			\multicolumn{1}{c|}{\textbf{*EGEUNET}}                 & \multirow{-8}{*}{\textbf{CNN-based}}                                                  & \textbf{0.927}          & \textbf{0.869}          & \textbf{4.146}            & \textbf{0.707}           & \multirow{-2}{*}{{\color[HTML]{FE0000} \textbf{+3.64\%}}} & \textbf{0.878}          & \multicolumn{1}{c|}{\textbf{0.805}}          & \textbf{0.910}          & \multicolumn{1}{c|}{\textbf{0.854}}          & \textbf{0.771}          & \multicolumn{1}{c|}{\textbf{0.681}}          & \textbf{0.787}          & \multicolumn{1}{c|}{\textbf{0.700}}          & \textbf{0.906}          & \multicolumn{1}{c|}{\textbf{0.847}}          & \multirow{-2}{*}{{\color[HTML]{FE0000} \textbf{+8.58\%}}}  \\ \midrule
			\multicolumn{1}{c|}{Convformer}                        &                                                                                       & 0.905                   & 0.835                   & 7.192                     & 1.143                    & {\color[HTML]{FE0000} }                                   & 0.867                   & \multicolumn{1}{c|}{0.792}                   & 0.813                   & \multicolumn{1}{c|}{0.747}                   & 0.722                   & \multicolumn{1}{c|}{0.637}                   & 0.664                   & \multicolumn{1}{c|}{0.588}                   & 0.868                   & \multicolumn{1}{c|}{0.794}                   & {\color[HTML]{FE0000} }                                    \\
			\multicolumn{1}{c|}{\textbf{*Convformer}}              &                                                                                       & \textbf{0.912}          & \textbf{0.844}          & \textbf{4.831}            & \textbf{0.738}           & \multirow{-2}{*}{{\color[HTML]{FE0000} \textbf{+0.83\%}}} & \textbf{0.880}          & \multicolumn{1}{c|}{\textbf{0.805}}          & \textbf{0.920}          & \multicolumn{1}{c|}{\textbf{0.869}}          & \textbf{0.783}          & \multicolumn{1}{c|}{\textbf{0.695}}          & \textbf{0.783}          & \multicolumn{1}{c|}{\textbf{0.697}}          & \textbf{0.892}          & \multicolumn{1}{c|}{\textbf{0.822}}          & \multirow{-2}{*}{{\color[HTML]{FE0000} \textbf{+8.24\%}}}  \\ \cmidrule(r){1-1} \cmidrule(l){3-18} 
			\multicolumn{1}{c|}{TransFuse}                         &                                                                                       & 0.896                   & 0.823                   & 7.338                     & 1.223                    & {\color[HTML]{FE0000} }                                   & 0.869                   & \multicolumn{1}{c|}{0.792}                   & 0.885                   & \multicolumn{1}{c|}{0.828}                   & 0.750                   & \multicolumn{1}{c|}{0.651}                   & 0.671                   & \multicolumn{1}{c|}{0.597}                   & 0.854                   & \multicolumn{1}{c|}{0.782}                   & {\color[HTML]{FE0000} }                                    \\
			\multicolumn{1}{c|}{\textbf{*TransFuse}}               &                                                                                       & \textbf{0.913}          & \textbf{0.849}          & \textbf{6.404}            & \textbf{1.106}           & \multirow{-2}{*}{{\color[HTML]{FE0000} \textbf{+1.93\%}}} & \textbf{0.891}          & \multicolumn{1}{c|}{\textbf{0.822}}          & \textbf{0.911}          & \multicolumn{1}{c|}{\textbf{0.854}}          & \textbf{0.791}          & \multicolumn{1}{c|}{\textbf{0.706}}          & \textbf{0.763}          & \multicolumn{1}{c|}{\textbf{0.676}}          & \textbf{0.904}          & \multicolumn{1}{c|}{\textbf{0.844}}          & \multirow{-2}{*}{{\color[HTML]{FE0000} \textbf{+5.73\%}}}  \\ \cmidrule(r){1-1} \cmidrule(l){3-18} 
			\multicolumn{1}{c|}{Xboundformer}                      &                                                                                       & 0.917                   & 0.851                   & 4.588                     & 0.695                    & {\color[HTML]{FE0000} }                                   & 0.857                   & \multicolumn{1}{c|}{0.786}                   & 0.824                   & \multicolumn{1}{c|}{0.757}                   & 0.777                   & \multicolumn{1}{c|}{0.697}                   & 0.727                   & \multicolumn{1}{c|}{0.651}                   & 0.895                   & \multicolumn{1}{c|}{0.836}                   & {\color[HTML]{FE0000} }                                    \\
			\multicolumn{1}{c|}{\textbf{*Xboundformer}}            & \multirow{-6}{*}{\textbf{\begin{tabular}[c]{@{}c@{}}CNN-\\ Transformer\\hybrid\end{tabular}}} & \textbf{0.936}          & \textbf{0.884}          & \textbf{2.985}            & \textbf{0.578}           & \multirow{-2}{*}{{\color[HTML]{FE0000} \textbf{+2.12\%}}} & \textbf{0.900}          & \multicolumn{1}{c|}{\textbf{0.835}}          & \textbf{0.913}          & \multicolumn{1}{c|}{\textbf{0.865}}          & \textbf{0.793}          & \multicolumn{1}{c|}{\textbf{0.704}}          & \textbf{0.804}          & \multicolumn{1}{c|}{\textbf{0.718}}          & \textbf{0.912}          & \multicolumn{1}{c|}{\textbf{0.857}}          & \multirow{-2}{*}{{\color[HTML]{FE0000} \textbf{+5.94\%}}}  \\ \midrule
			\multicolumn{1}{c|}{SwinUmamba}                        &                                                                                       & 0.906                   & 0.836                   & 5.640                     & 0.899                    & {\color[HTML]{FE0000} }                                   & 0.853                   & \multicolumn{1}{c|}{0.782}                   & 0.876                   & \multicolumn{1}{c|}{0.820}                   & 0.715                   & \multicolumn{1}{c|}{0.626}                   & 0.667                   & \multicolumn{1}{c|}{0.589}                   & 0.861                   & \multicolumn{1}{c|}{0.789}                   & {\color[HTML]{FE0000} }                                    \\
			\multicolumn{1}{c|}{\textbf{*SwinUmamba}}              & \multirow{-2}{*}{\textbf{Mamba}}                                                      & \textbf{0.922}          & \textbf{0.860}          & \textbf{3.845}            & \textbf{0.690}           & \multirow{-2}{*}{{\color[HTML]{FE0000} \textbf{+1.69\%}}} & \textbf{0.879}          & \multicolumn{1}{c|}{\textbf{0.802}}          & \textbf{0.916}          & \multicolumn{1}{c|}{\textbf{0.863}}          & \textbf{0.794}          & \multicolumn{1}{c|}{\textbf{0.692}}          & \textbf{0.785}          & \multicolumn{1}{c|}{\textbf{0.701}}          & \textbf{0.905}          & \multicolumn{1}{c|}{\textbf{0.846}}          & \multirow{-2}{*}{{\color[HTML]{FE0000} \textbf{+7.75\%}}}  \\ \midrule
			\multicolumn{2}{c|}{\textbf{Average Growth Rate}}                                                                                               & \multicolumn{5}{c|}{{\color[HTML]{FE0000} \textbf{+2.37\%}}}                                                                                                         & \multicolumn{11}{c}{{\color[HTML]{FE0000} \textbf{+7.16\%}}}                                                                                                                                                                                                                                                                                                                                                                            \\ \bottomrule
	\end{tabular}}
	\caption{Evaluation of the proposed DALE on various advanced models using ISIC2016\&ph2 and Polyp datasets. }
	\label{tab:2}
\end{table*}

\section{Experiments}
\label{exper}
\subsection{Experimental Setup}
\mypara{Datasets.} We evaluate the proposed DALE on four public datasets.
 \textbf{MosMedData+} \cite{morozov2020mosmeddata}:  This dataset contains 2729 CT slices of lung infections, split into 2183 training images, 273 validation images, and 273 test images. All images are cropped to $224 \times 224$, with data augmentation applied using random scaling (10\% probability) \cite{li2024lvit}.
\textbf{ISIC \& ph2} \cite{gutman2016skin, mendoncca2013ph}: The ISIC-2016 dataset contains 900 training samples and 379 validation samples, while the ph2 dataset includes 200 samples. We train the model on ISIC-2016 and test it on ph2 to evaluate segmentation accuracy and generalization \cite{lee2020structure, wang2023xbound}.
 \textbf{Polyp}: Five public polyp datasets are used: Kvasir \cite{jha2020kvasir}, CVC-ClinicDB \cite{tajbakhsh2015automated}, CVC-ColonDB \cite{bernal2015wm}, EndoScene \cite{vazquez2017benchmark}, and ETIS \cite{silva2014toward}. Following the split and training settings in \cite{fan2020pranet, zhang2021transfuse}, we use 1450 training images from Kvasir and CVC-ClinicDB, and 798 testing images from all five datasets.
\textbf{Synapse}: A multi-organ segmentation dataset consisting of 30 abdominal CT scans (each scan includes 85-198 slices) for the segmentation of 8 abdominal structures. The dataset is split into 18 training scans (2212 images) and 12 validation scans (1567 images) \cite{chen2021transunet}.

\noindent\mypara{Evaluation metrics.}  The segmentation performance is evaluated using metrics such as Dice coefficient (Dice), mean Intersection over Union (mIoU), 95\% Hausdorff Distance (95HD), and Average Surface Distance (ASD).

\noindent\mypara{Implementation details.} All experiments are conducted in PyTorch \cite{paszke2019pytorch} on an NVIDIA L40 GPU with 48 GB of memory. The models are optimized by the Adam optimizer \cite{kingma2014adam}, with a learning rate of 3e-4, batch size of 24, $\omega = 0.1$, $K = 1$, $\tau = 0.9$, $\alpha = 0.05$, and $h \times w = 16 \times 16$.
Experimental results represent the average of five trials, with significance confirmed by a paired T-test at 0.05 demonstrating the stability of the proposed training paradigm. 

\noindent\mypara{Baselines.}  Our method replaces traditional training strategies for several models, including CNN-based architectures (U-Net \cite{ronneberger2015u}, U-Net++ \cite{zhou2018unet++}, EGEUNET \cite{ruan2023ege}, EMCAD \cite{rahman2024emcad}); CNN-Transformer hybrid models (ConvFormer \cite{lin2023convformer}, TransFuse \cite{zhang2021transfuse}, XboundFormer \cite{wang2023xbound}); Mamba-based models (SwinUmamba \cite{liu2024swin}); and text-image multimodal models (LanGuide \cite{zhong2023ariadne}).
The compared methods include BECO \cite{rong2023boundary}, GCE \cite{zhang2018generalized},  MW-Net \cite{shu2019meta},  CIRL \cite{chen2024mind}. Our code is available at \textit{\url{https://github.com/lxFang2022/DALE}}.

%

\subsection{Comparison with State-of-the-art Methods}
\noindent\mypara{Results on different model types.} 
\cref{tab:1,tab:2} shows that DALE achieves superior performance across all lesion segmentation models and tasks compared to the traditional training paradigm.
On the MosMedData+ dataset, which contains many fuzzy regions (\cref{tab:1}), DALE achieves an average improvement of 5.33\%, demonstrating its ability to mitigate noise and calibrate unstable features in fuzzy regions.
\cref{tab:2} reports average performance improvements of 2.37\% and 7.16\% on the ISIC2017\&ph2 dataset and the Polyp dataset, respectively. These results demonstrate the method's ability to extract well-generalized features, enhancing model adaptability across datasets from different sources.
Beyond lesion segmentation, DALE is also evaluated in the multi-organ segmentation task, with a performance improvement of 2.27\% (Tab.1 in the Appendix) indicating its effectiveness.  

\begin{table}[]
	\small
	\scalebox{0.68}{
		\begin{tabular}{@{}c|cccc|cccc@{}}
			\toprule
			& \multicolumn{4}{c|}{\textbf{ISIC2016\&ph2}}                                                                               & \multicolumn{4}{c}{\textbf{MosMedData+}}                                                                                   \\ \cmidrule(l){2-9} 
			\multirow{-2}{*}{\textbf{Method}} & \textbf{Dice$\uparrow$}      & \textbf{mIoU$\uparrow$}      & \textbf{95HD$\downarrow$}    & \textbf{ASD$\downarrow$}     & \textbf{Dice$\uparrow$}      & \textbf{mIoU$\uparrow$}      & \textbf{95HD$\downarrow$}     & \textbf{ASD$\downarrow$}     \\ \midrule
			\textbf{Xboundformer}             & 0.917                        & 0.851                        & 4.588                        & {\color[HTML]{3531FF} 0.695} & 0.679                        & 0.543                        & 26.056                        & 6.257                        \\
			\textbf{BECO}                     & 0.881                        & 0.804                        & 10.424                       & 2.203                        & 0.675                        & 0.533                        & 23.435                        & {\color[HTML]{3531FF} 4.805} \\
			\textbf{GCE}                      & 0.882                        & 0.805                        & 8.140                        & 1.885                        & 0.689                        & 0.552                        & {\color[HTML]{3531FF} 21.534} & 4.974                        \\
			\textbf{MW-Net}                   & 0.919                        & 0.854                        & 5.470                        & 0.857                        & 0.693                        & 0.544                        & 25.781                        & 6.231                        \\
			\textbf{CIRL}                     & {\color[HTML]{3531FF} 0.928} & {\color[HTML]{3531FF} 0.870} & {\color[HTML]{3531FF} 4.097} & 0.750                        & {\color[HTML]{3531FF} 0.706} & {\color[HTML]{3531FF} 0.560} & 22.349                        & 4.987                        \\
			\textbf{DALE}                     & \textbf{0.936}               & \textbf{0.884}               & \textbf{3.224}               & \textbf{0.595}               & \textbf{0.709}               & \textbf{0.572}               & \textbf{20.534}               & \textbf{4.674}               \\ \bottomrule
	\end{tabular}}
	\caption{Quantitative comparison of state-of-the-art methods. Best results are in bold, second-best in blue.}
	\label{tab:3}
\end{table}
\begin{figure}[h]
	\centering
	\includegraphics[width=0.8\linewidth]{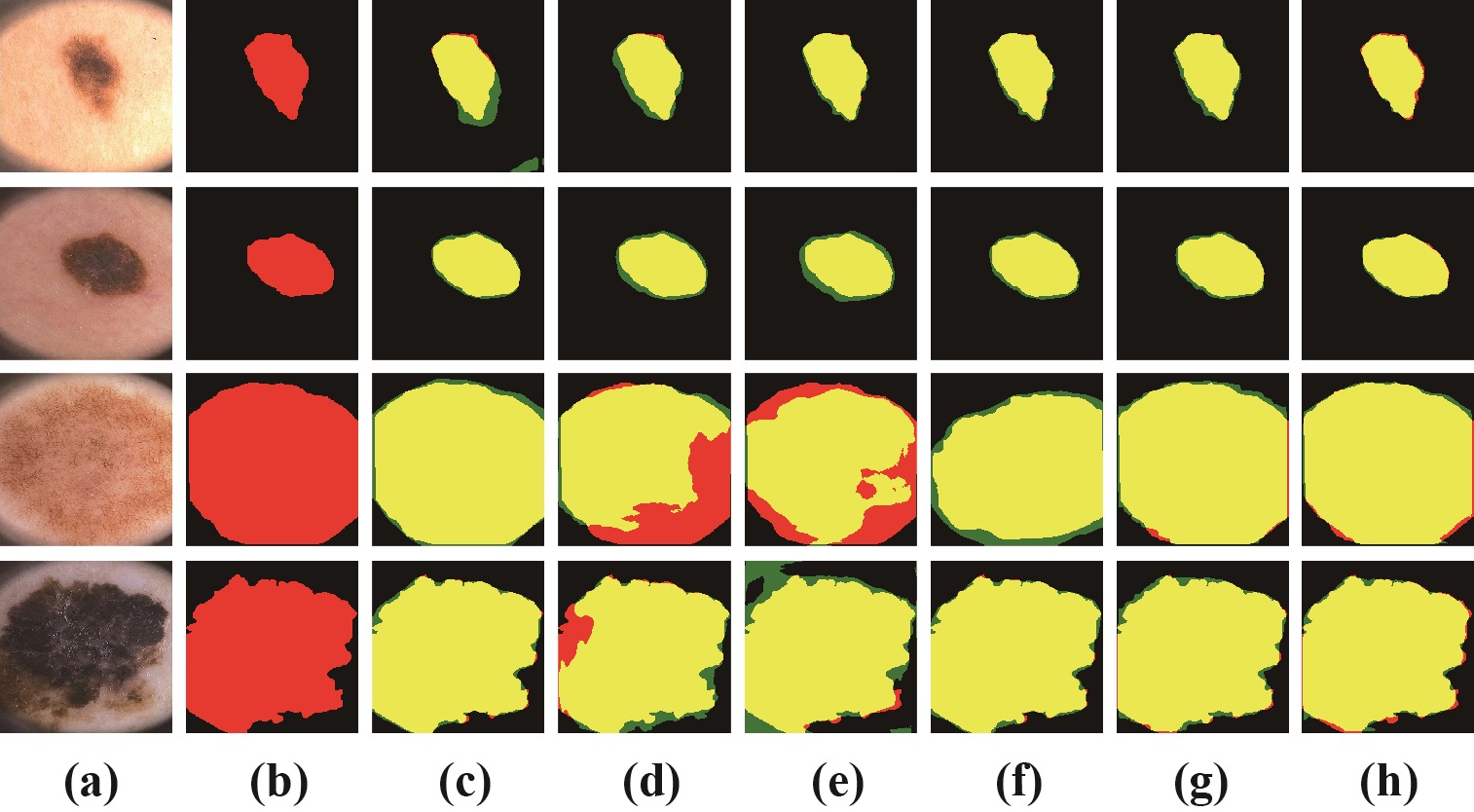}
	
	\caption{Visual comparison of different state-of-the-art methods on the ph2 datasets. (a) Images; (b) Ground truth labels; (c) Xboundformer; (d) BECO; (e) GCE; (f) MW-Net; (g) CIRL; (h) DALE. Red, green, yellow: ground truth, prediction, overlap.}
	\label{fig:7}
\end{figure}

\begin{figure}[b]
	\centering
	\begin{subfigure}{0.49\linewidth}
		\includegraphics[width=\textwidth]{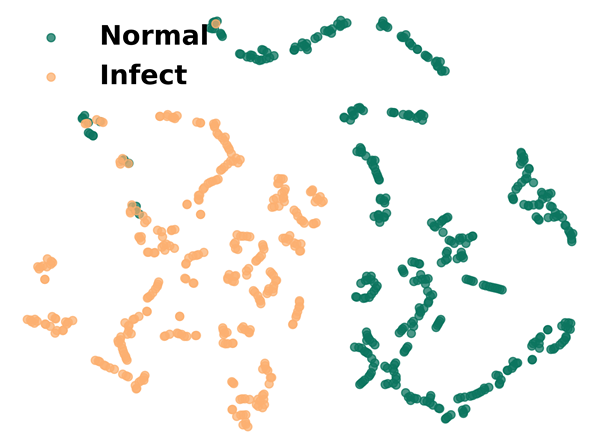}
		\caption{DALE}
		\label{fig:4-a}
	\end{subfigure}
	\hfill
	\begin{subfigure}{0.49\linewidth}
		\includegraphics[width=\textwidth]{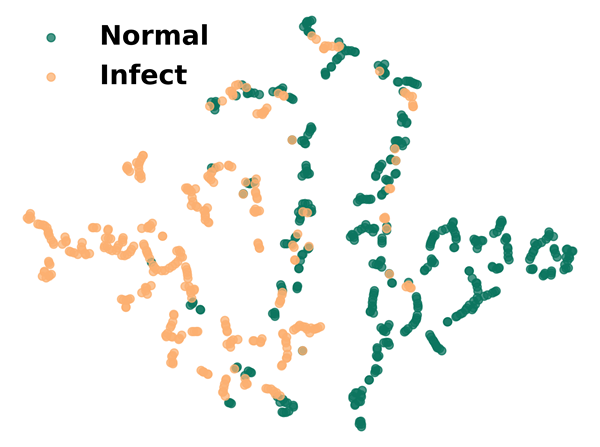} 
		\caption{CIRL}
		\label{fig:4-b}
	\end{subfigure}
	\caption{T-SNE visualization of features in fuzzy regions.} 
	\label{fig:4}  
\end{figure}

\begin{figure*}[t!]
	\centering
	\begin{subfigure}{0.2\linewidth}
		\centering
		\includegraphics[scale=0.55]{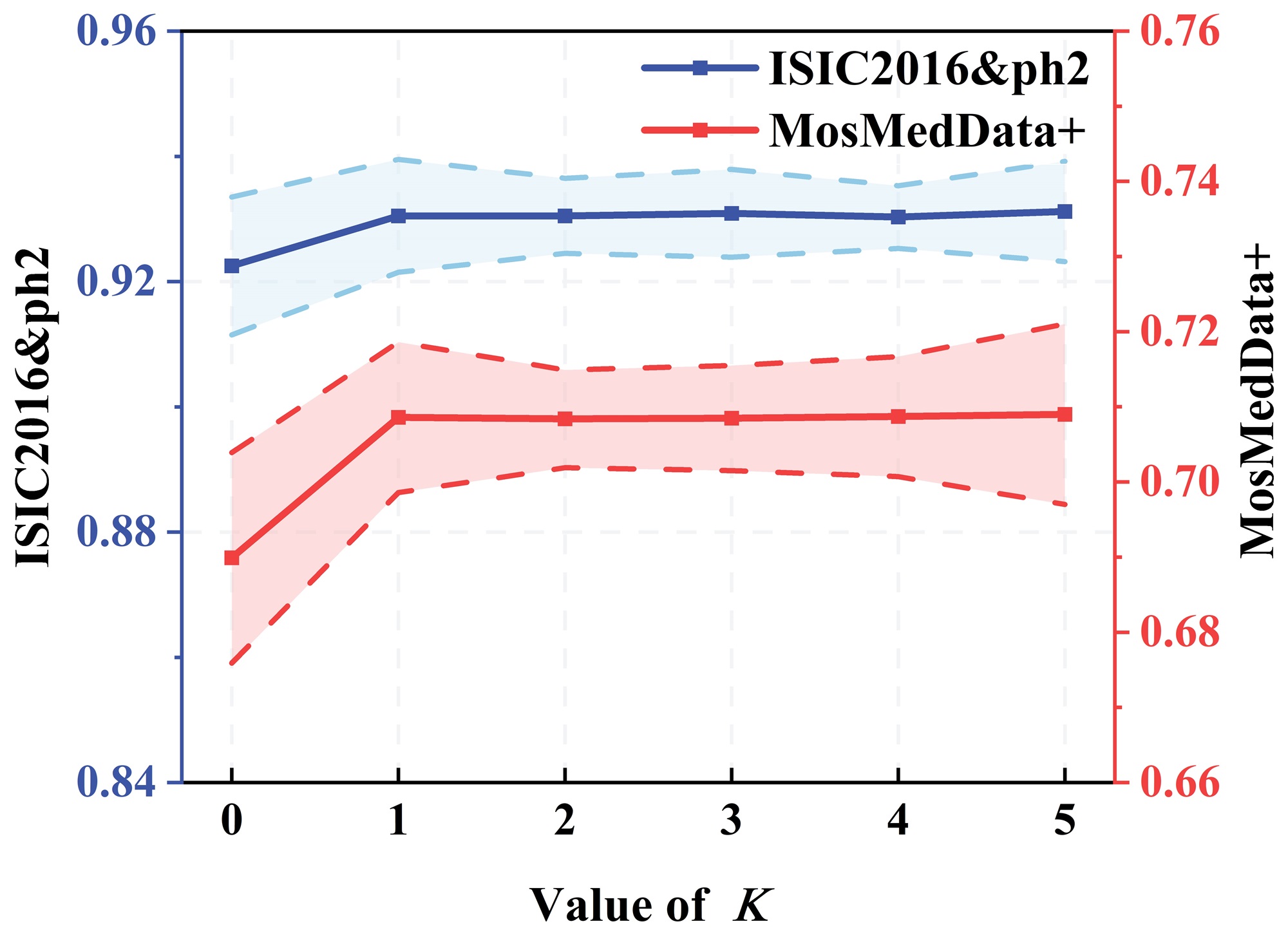}
		\caption{Optimization iterations $K$.}
		\label{fig:6-a}
	\end{subfigure}
	\hfill
	\begin{subfigure}{0.2\linewidth}
		\centering
		\includegraphics[scale=0.18]{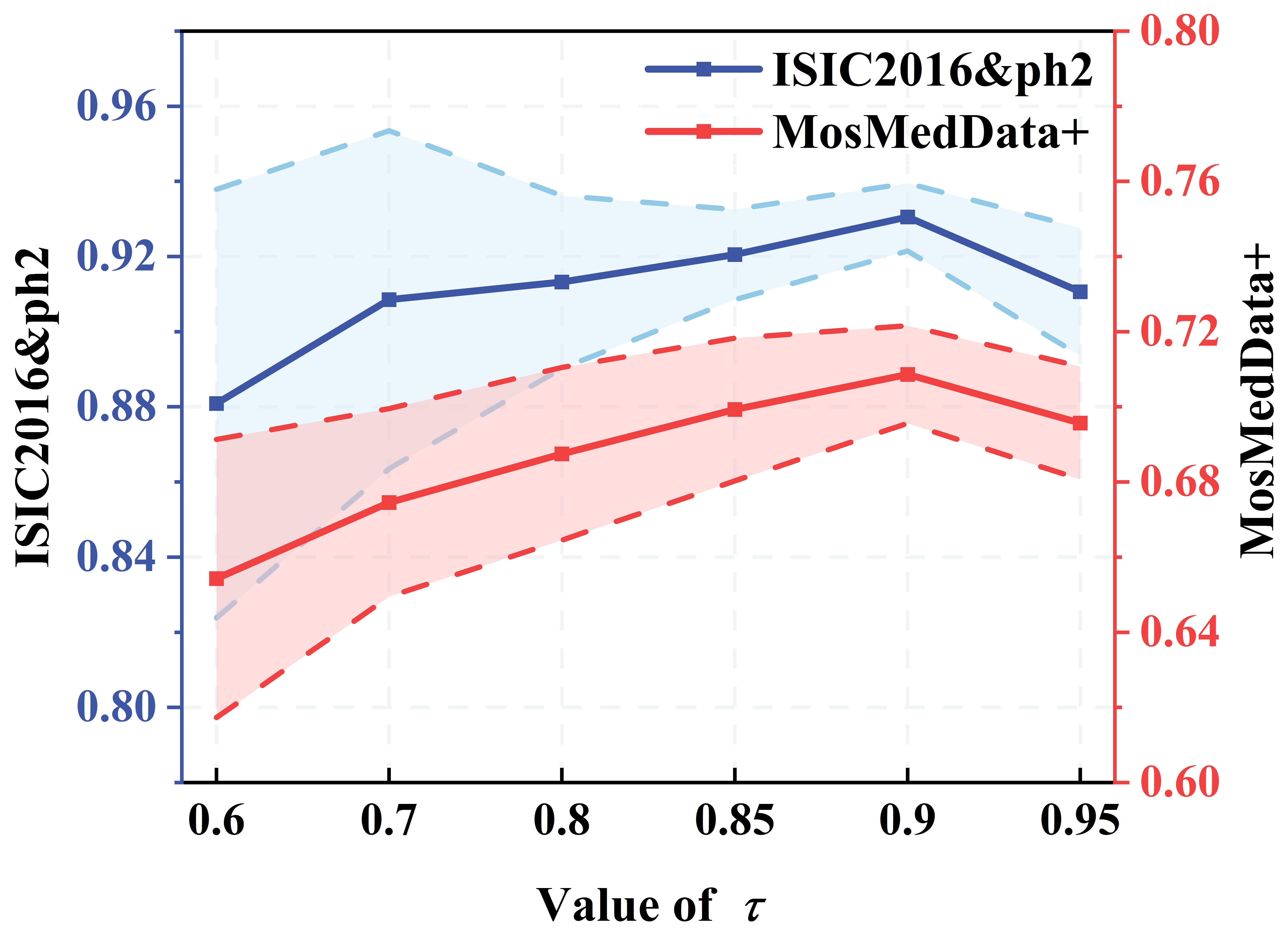} 
		\caption{Soft threshold $\tau$.}
		\label{fig:6-b}
	\end{subfigure}	\hfill
	\begin{subfigure}{0.2\linewidth}
		\centering
		\includegraphics[scale=0.55]{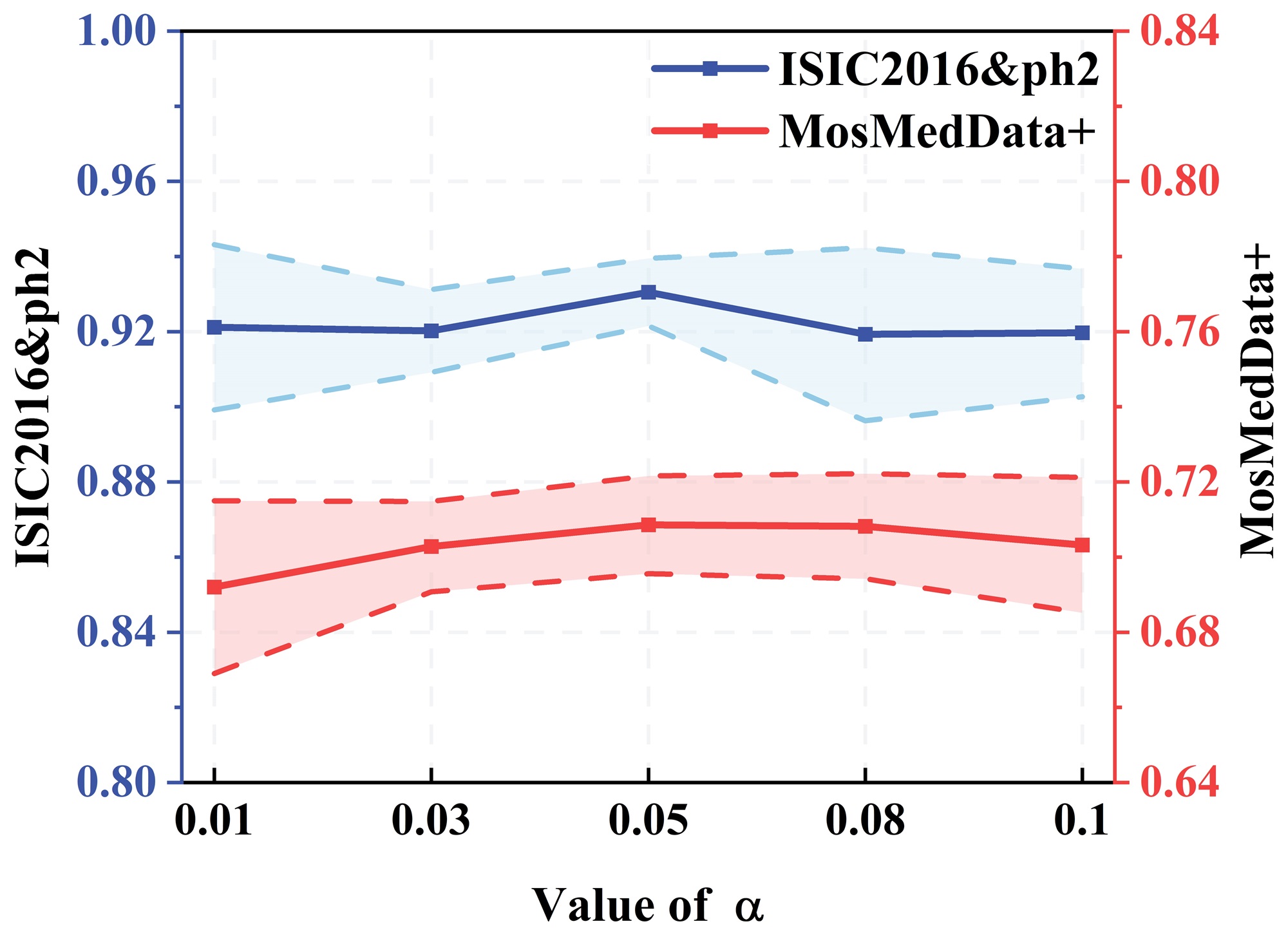}
		\caption{Loss weight $\alpha$ for $\mathcal{L}_W$.}
		\label{fig:6-c}
	\end{subfigure}
	\hfill
	\begin{subfigure}{0.2\linewidth}
		\centering
		\includegraphics[scale=0.55]{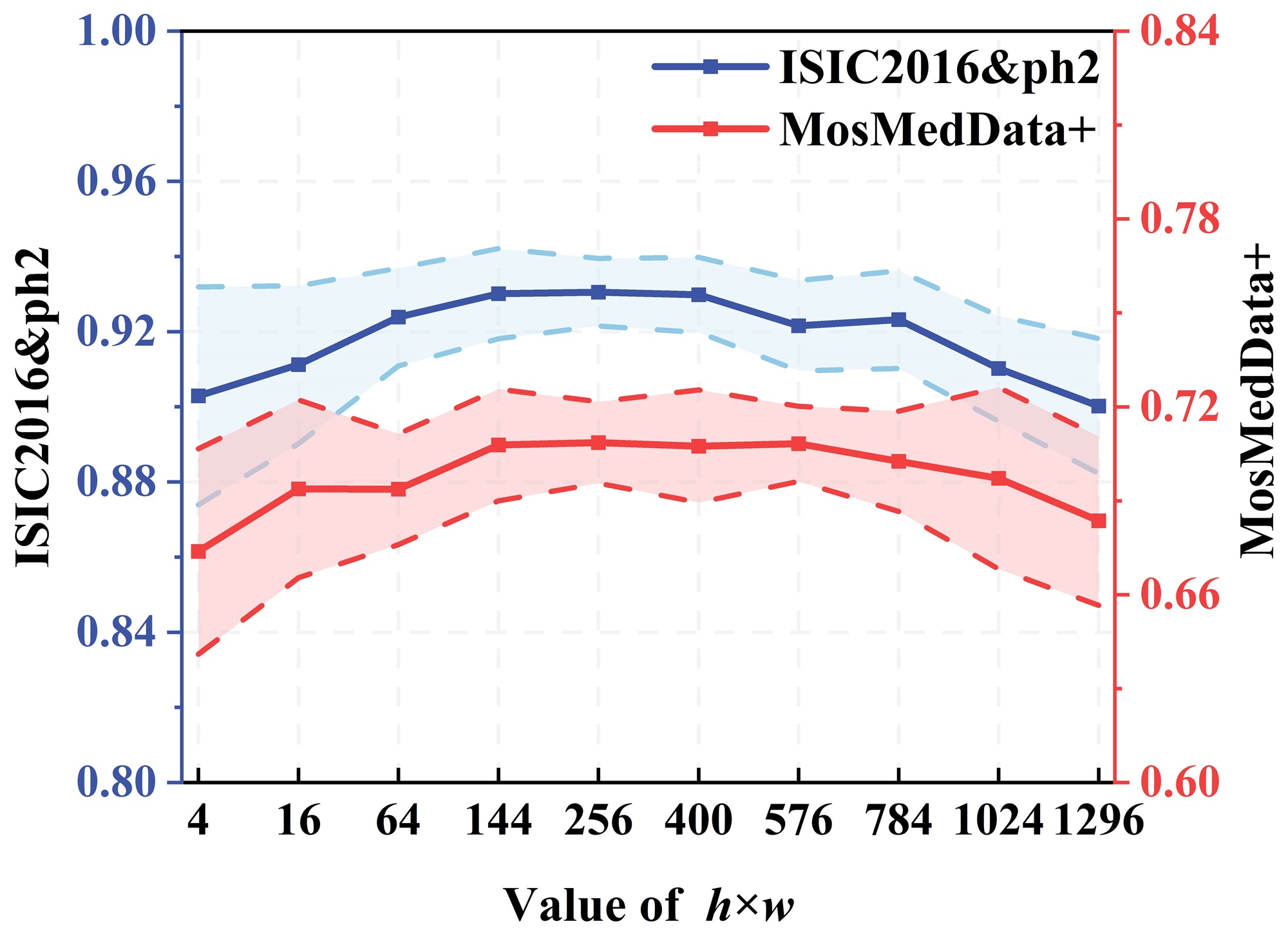} 
		\caption{Patch size $h \times w$.}
		\label{fig:6-d}
	\end{subfigure}
	\caption{Sensitivity analysis of parameters $K$, $\tau$, $\alpha$ and patch size $h \times w$ for DALE applied on the Xboundformer model.} 
	\label{fig:6}  
\end{figure*}
\begin{figure}[t]
	\centering
	\begin{subfigure}{0.49\linewidth}
		\includegraphics[width=\textwidth]{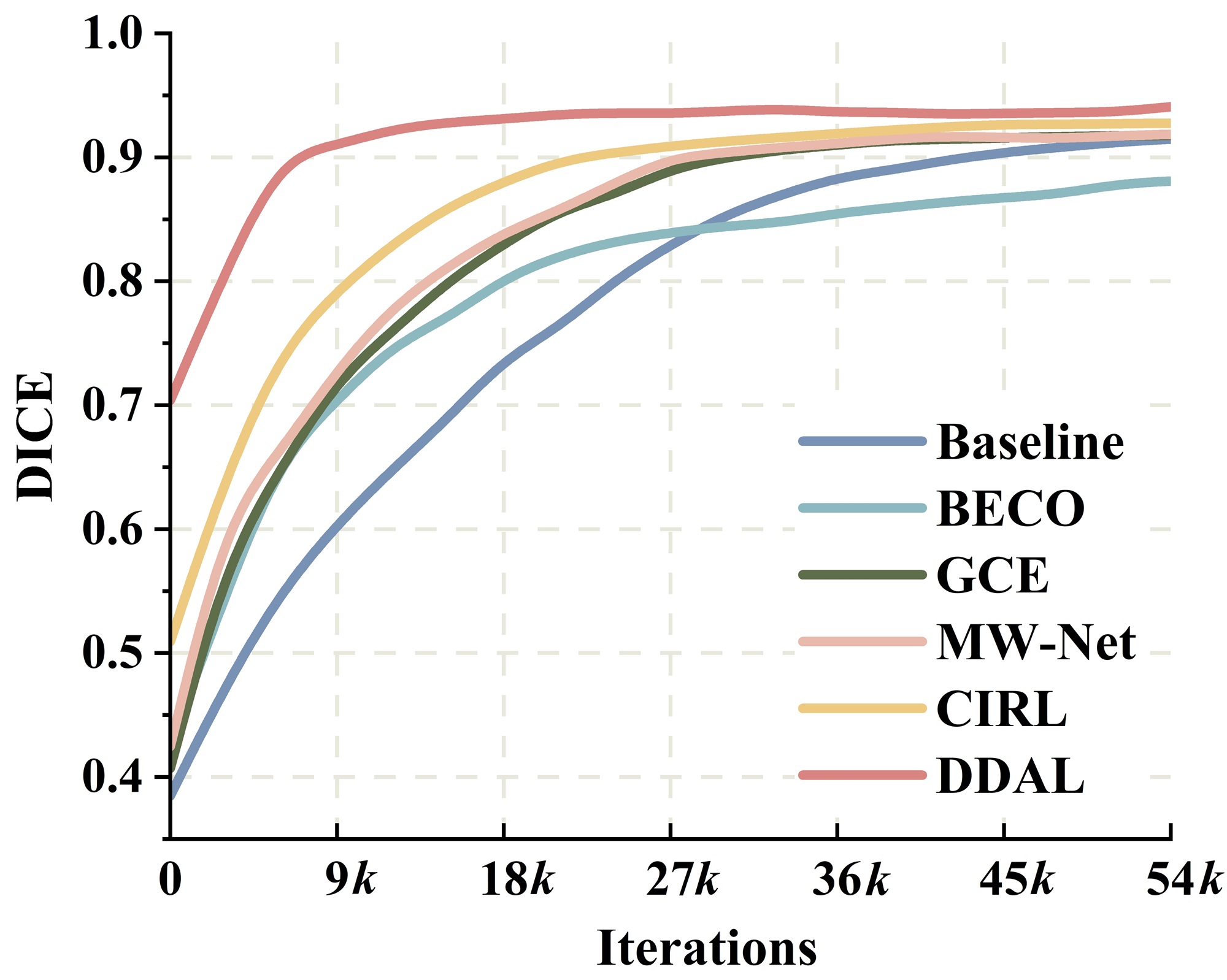}
		\caption{Convergence of DALE.}
		\label{fig:5-a}
	\end{subfigure}
	\hfill
	\begin{subfigure}{0.49\linewidth}
		\includegraphics[width=\textwidth]{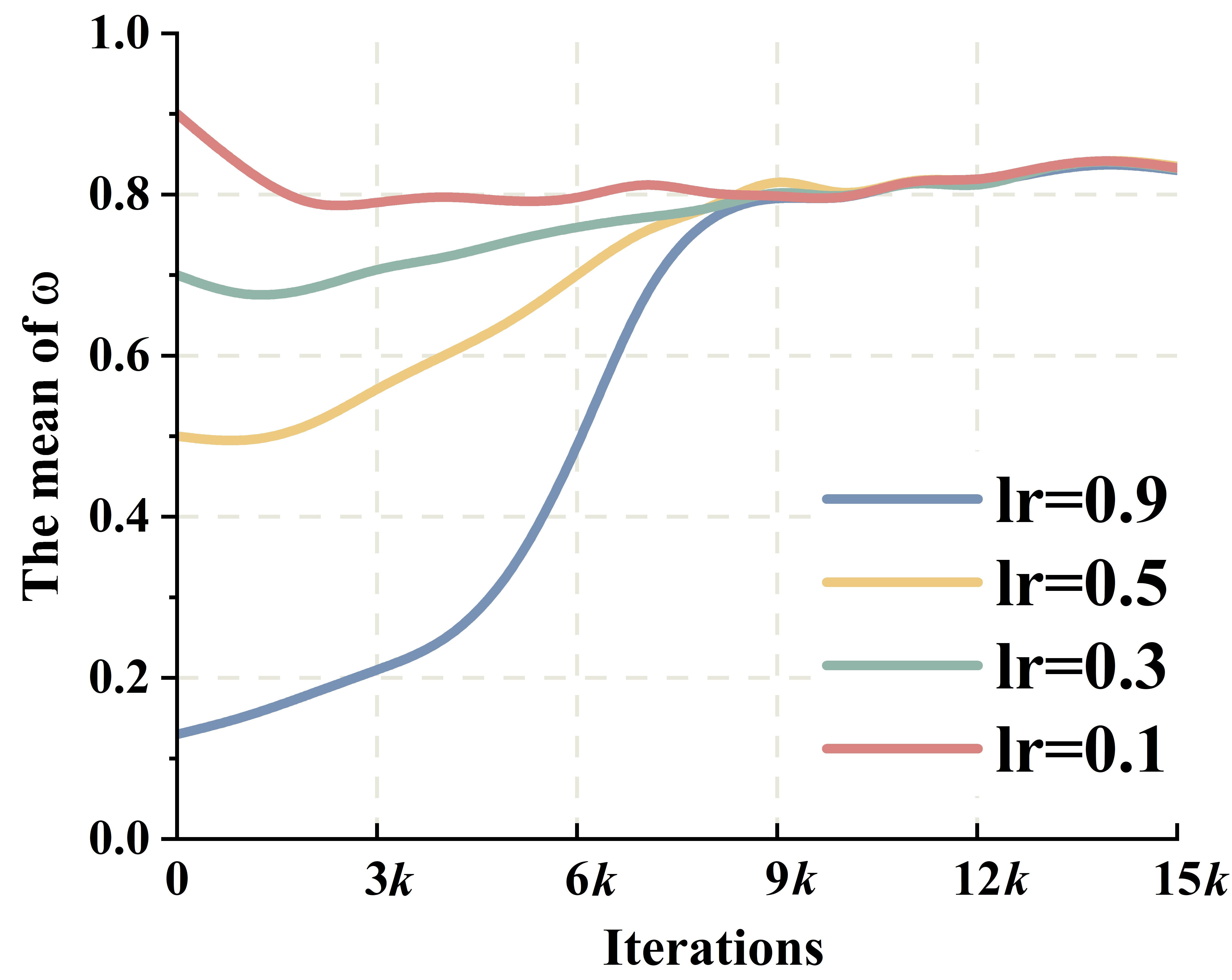} 
		\caption{Learning rate impact on $\omega$.}
		\label{fig:5-b}
	\end{subfigure}
	\caption{Convergence of DALE and learning rate impact on $\omega$.} 
	\label{fig:5}  
\end{figure}

\noindent\mypara{Results compared with advanced methods.}
\cref{tab:3,fig:7} shows the results of all competing methods on the ISIC2017\&ph2 and MosMedData+ datasets. All evaluation metrics show that our method performs best.
BECO \cite{rong2023boundary} shows a performance drop, primarily due to replacing all fuzzy region samples with pseudo-labels, which disrupts many originally clean samples.
CIRL \cite{chen2024mind} achieves some improvement by separating fuzzy and non-fuzzy regions, employing unsupervised learning in fuzzy regions to mitigate noisy label interference. 
Note that compared to CIRL, our approach further reduces the 95HD by 21.3\% and the ASD by 20.67\%, indicating superior performance in segmenting unclear boundaries. Unlike CIRL, which uses unsupervised learning for fuzzy regions, our method filters out noisy labels in these regions, reducing their negative impact on model supervision. \cref{fig:4} shows that DALE enables the model to learn discriminative features in fuzzy regions more effectively than CIRL. Moreover, \cref{fig:5-a} demonstrates that DALE achieves a faster convergence rate and higher stability.

\subsection{Ablation Studies}

\noindent\mypara{Effectiveness of components.}
\cref{tab:4} shows that each component enhances the overall performance of DALE significantly. In the AE\&ER soft threshold method, average entropy and edge pixel ratio are both effective in identifying fuzzy regions, and their combined use maximizes performance. In fuzzy regions, $\omega$ improves segmentation performance in fuzzy regions significantly, demonstrating the effectiveness of loss consistency-based collaborative optimization. The proposed unstable representation calibration method also improves fuzzy region representation learning abilities and provides great improvements.

\noindent\mypara{Ablation of hyper-parameters.}
\cref{fig:5-b,fig:6} presents the results of hyper-parameter sensitivity experiments, including the learning rate of parameter $\omega$, the number of optimization iterations $K$ for parameter $\omega$, the soft threshold $\tau$, the distribution calibration loss weight $\alpha$, and the patch size $h \times w$. \cref{fig:5-b} indicates that different learning rates affect the initial convergence speed of the $\omega$ optimization, but have little effect on the final performance. 
\begin{table}[]
	\centering
	\small
	\scalebox{0.75}{
		\begin{tabular}{@{}cccccccc@{}}
			\toprule
			&                               &                                     &                                  & \multicolumn{4}{c}{\textbf{ISIC2016\&ph2}}                                                                                    \\ \cmidrule(l){5-8} 
			\multirow{-2}{*}{\textbf{AE}} & \multirow{-2}{*}{\textbf{ER}} & \multirow{-2}{*}{\textbf{$\bm{\omega}$}} & \multirow{-2}{*}{\textbf{\bm{$\mathcal{L}_W$}}} & \textbf{Dice$\uparrow$}       & \textbf{mIoU$\uparrow$}       & \textbf{95HD$\downarrow$}     & \textbf{ASD$\downarrow$}      \\ \midrule
			&                               &                                     &                                  & 0.9165                        & 0.8508                        & 4.5883                        & 0.6951                        \\
			\checkmark                    &                               &                                     &                                  & 0.9198                        & 0.8574                        & 4.1038                        & 0.6856                        \\
			& \checkmark                    &                                     &                                  & 0.9220                        & 0.8612                        & 4.4498                        & 0.7398                        \\
			\checkmark                    & \checkmark                    &                                     &                                  & 0.9225                        & 0.8616                        & 4.1411                        & 0.6260                        \\
			\checkmark                    &                               & \checkmark                          &                                  & 0.9258                        & 0.8667                        & 4.5589                        & {\color[HTML]{3531FF} 0.5816} \\
			& \checkmark                    & \checkmark                          &                                  & 0.9266                        & 0.8685                        & 4.1091                        & 0.6211                        \\
			\checkmark                    & \checkmark                    & \checkmark                          &                                  & {\color[HTML]{3531FF} 0.9276} & {\color[HTML]{3531FF} 0.8709} & {\color[HTML]{3531FF} 4.0785} & 0.6192                        \\
			\checkmark                    & \checkmark                    & \checkmark                          & \checkmark                       & \textbf{0.9359}               & \textbf{0.8840}               & \textbf{2.9852}               & \textbf{0.5780}               \\ \bottomrule
		\end{tabular}
	}
	\caption{Ablation results of DALE components.}
	\label{tab:4}
\end{table}
\begin{table}[]
		\small
	\scalebox{0.63}{
	\begin{tabular}{@{}c|cc|cc|cc@{}}
		\toprule
		& \textbf{Xboundformer} & \textbf{+DALE} & \textbf{EGEUNET} & \textbf{+DALE} & \textbf{SwinUmamba} & \textbf{+DALE} \\ \midrule
		\textbf{Iter. Time (s)} & 0.15                  & 0.51           & 0.13             & 0.19           & 0.16                & 0.68           \\
		\textbf{Iter. Nums}     & 124200                & 14400          & 88200            & 20700          & 140400              & 22500          \\
		\textbf{Total Time (h)} & 5.06                  & 2.55           & 3.19             & 1.09           & 6.24                & 4.25           \\ \bottomrule
	\end{tabular}}
		\caption{Time analysis.}
	\label{tab:5}
\end{table}
\subsection{Limitation and Discussion}
DALE takes a long time to run each iteration, but this is due to the gradient backpropagation process being performed twice per iteration for non-fuzzy and fuzzy sets. However, since DALE's alternate learning strategy accelerates convergence and reduces training rounds (\cref{fig:5-a}), total training time is reduced (\cref{tab:5}).

%% file: flx/Conclusion.tex
\section{Conclusion}
We present a novel learning paradigm called data-driven alternating learning (DALE). This approach optimizes model training to address challenges like unclear boundaries and label ambiguities, enabling stable lesion segmentation. The core idea is to guide fuzzy R-set learning using reliable non-fuzzy R-sets. During fuzzy R-set training, the proposed loss consistency-based collaborative optimization adaptively adjusts the influence of confidence labels to reduce errors. 
Moreover, the proposed unstable representation calibration strategy based on distribution alignment enhances the model's ability to learn stable and discriminative features in fuzzy regions.
Notably, the DALE paradigm is a model-agnostic training approach that can be easily applied to various segmentation models. 
Extensive experiments on multiple benchmarks and model backbones demonstrate DALE's effectiveness in handling fuzzy regions in medical image segmentation. 